\documentclass[preprint,12pt]{elsarticle}

\usepackage{amssymb}
\usepackage{amsmath}

\usepackage{multirow}%
\usepackage{caption}
\usepackage{booktabs}
\usepackage{float}

\journal{Nuclear Physics B}

\begin{document}

\begin{frontmatter}



\title{MonoPartNeRF:Human Reconstruction from Monocular Video via Part-Based Neural Radiance Fields}

\author[1,2,3]{Yao Lu}
\author[1,2,3]{Jiawei Li}
\author[1,2,4]{Ming Jiang}

\affiliation[1]{organization={The School of Computer Science and Information Security, Guilin University of Electronic Technology},
            addressline={},
            city={Guilin}, 
            postcode={541004}, 
            state={Guangxi}, 
            country={China}}

\affiliation[2]{organization={Metaverse Application Engineering Center},
            addressline={},
            city={Nanning}, 
            postcode={530000}, 
            state={Guangxi}, 
            country={China}}

\affiliation[3]{These authors contributed equally to this work}

\affiliation[4]{Corresponding author: mjiang@guet.edu.cn}

\begin{abstract}
In recent years, Neural Radiance Fields (NeRF) have achieved remarkable progress in dynamic human reconstruction and rendering. Part-based rendering paradigms, guided by human segmentation, allow for flexible parameter allocation based on structural complexity, thereby enhancing representational efficiency. However, existing methods still struggle with complex pose variations, often producing unnatural transitions at part boundaries and failing to reconstruct occluded regions accurately in monocular settings. We propose MonoPartNeRF, a novel framework for monocular dynamic human rendering that ensures smooth transitions and robust occlusion recovery. First, we build a bidirectional deformation model that combines rigid and non-rigid transformations to establish a continuous, reversible mapping between observation and canonical spaces. Sampling points are projected into a parameterized surface-time space (u, v, t) to better capture non-rigid motion. A consistency loss further suppresses deformation-induced artifacts and discontinuities. We introduce a part-based pose embedding mechanism that decomposes global pose vectors into local joint embeddings based on body regions. This is combined with keyframe pose retrieval and interpolation, along three orthogonal directions, to guide pose-aware feature sampling. A learnable appearance code is integrated via attention to model dynamic texture changes effectively. Experiments on the ZJU-MoCap and MonoCap datasets demonstrate that our method significantly outperforms prior approaches under complex pose and occlusion conditions, achieving superior joint alignment, texture fidelity, and structural continuity.
\end{abstract}




\end{frontmatter}



\section{Introduction}\label{sec1}

Free-viewpoint human rendering holds significant potential in various fields such as virtual reality (VR), extended reality (XR), film production, gaming, and remote video conferencing. However, achieving high-quality, photorealistic dynamic human rendering remains challenging, especially under monocular viewpoints. Traditional methods often rely on multi-view images captured from dense camera arrays \cite{debevec2000acquiring,guo2019relightables} or precise depth information \cite{collet2015high,dou2016fusion4d}, which are both costly and labor-intensive to acquire. Furthermore, the rendering process must not only accurately capture complex joint movements but also handle non-rigid deformations from unseen viewpoints (such as clothing wrinkles), posing higher demands on human body modeling techniques.

The emergence of Neural Radiance Fields (NeRF) \cite{mildenhall2021nerf} has provided a new paradigm for scene representation, modeling the scene as a continuous five-dimensional function and demonstrating great potential in free-viewpoint rendering. Subsequent works have extended NeRF to dynamic scenes \cite{pons2021d,li2021neural}, enabling the handling of temporal data and the rendering of dynamic environments. These advancements have significantly improved the applicability of NeRF-based models in practice and offered new solutions for dynamic human modeling, motion capture, and virtual reality. However, generating realistic dynamic humans efficiently under monocular or sparse-view conditions still poses significant challenges.

To address dynamic human rendering under monocular or sparse viewpoints, some studies \cite{peng2021animatable,weng2022humannerf,yu2023monohuman,li2023posevocab} leverage NeRF-based approaches to enhance the quality of 3D human model generation by integrating human pose and motion details, thereby improving realism and refinement across varying viewpoints. Although these methods simplify the modeling process, they suffer from certain limitations. The structural and detail complexity varies significantly across different human body parts. Uniform density distributions constrain the network's ability to represent highly complex areas (such as joints, hands, and face) and lead to capacity waste in less complex regions, thereby reducing overall efficiency. Moreover, single-network architectures struggle with occlusions, resulting in shape distortions and texture artifacts due to missing visual information, which degrades rendering quality. Instant-NVR \cite{geng2023learning} addresses this by decomposing the human body into multiple parts for separate rendering and adaptively allocating network capacity according to the complexity of each part. While this method improves detail representation in complex regions and optimizes capacity allocation, it still faces limitations when handling extreme pose variations, leading to unnatural joint connections and difficulty maintaining plausible body shapes. In monocular settings, limited camera viewpoints cause parts of the body to be self-occluded, resulting in shape distortions or texture inaccuracies.

\begin{figure}[t]
	\centering
	\includegraphics[width=0.8\linewidth]{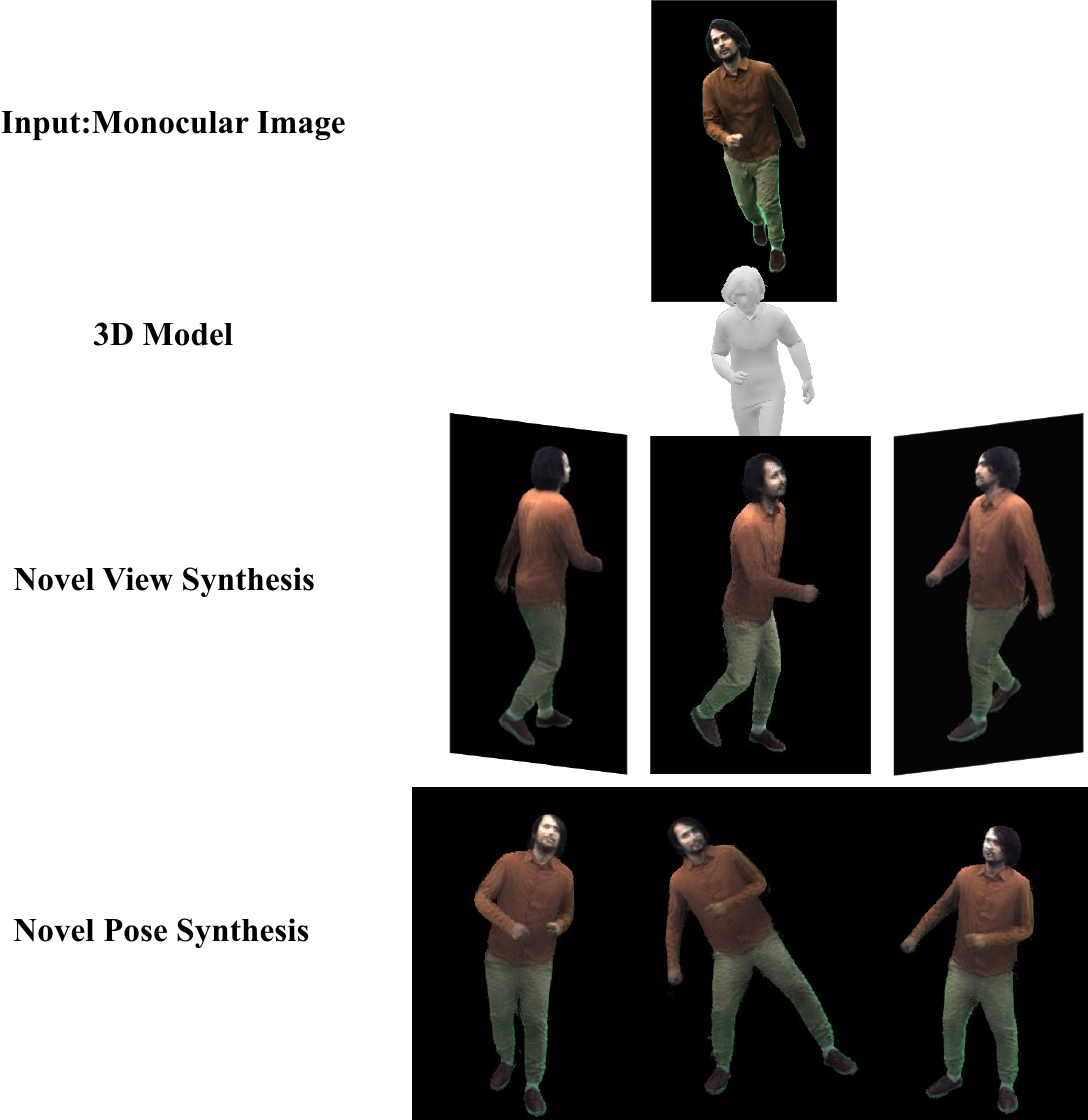}
	\caption{Our model is capable of generating realistic novel views from a monocular perspective.} 
	\label{fig:1}
\end{figure}
To tackle these challenges in monocular dynamic human rendering, we propose a novel rendering framework, \textbf{MonoPartNeRF}. This method decomposes the human body into multiple independent anatomical parts and introduces a bidirectional deformation mechanism to effectively model complex geometric changes caused by non-rigid motion. During the non-rigid deformation stage, MonoPartNeRF integrates UV-parameterized motion with temporal frame information and introduces a consistency loss to ensure accurate alignment of body parts into a unified canonical space under varying poses, thus achieving disentangled modeling of pose and geometry while maintaining natural and continuous joint connections. Furthermore, we design a part-aware pose embedding mechanism that decomposes the global pose representation into local joint rotations. For each query point, we employ similarity-based key-pose search and interpolation to achieve fine-grained pose feature alignment. Finally, the interpolated pose features are orthogonally projected into 3D space and fused with local appearance encoding to guide the neural rendering network in generating richly detailed images. Experiments demonstrate that MonoPartNeRF consistently produces stable and realistic rendering results under complex poses and occlusion conditions, significantly enhancing geometric fidelity and texture consistency. As shown in Fig.~\ref{fig:1}, our method achieves high-quality human rendering under both novel viewpoints and novel poses, highlighting its superior performance and generalization ability in dynamic human rendering tasks.

In summary, our main contributions are as follows:
\begin{itemize}
  \setlength{\leftskip}{1em}
  \item We propose \textbf{MonoPartNeRF}, a monocular free-viewpoint and pose rendering framework for dynamic humans, which generates structurally coherent and detail-rich novel views and poses without requiring multi-view or depth supervision, significantly improving generalization to unseen viewpoints and poses.
  \item We decompose the human body into multiple semantically consistent parts and introduce a bidirectional deformation mechanism within each part. By incorporating surface-time modeling with UV parameterization and temporal information during bidirectional non-rigid deformation, we achieve natural and continuous joint connections and geometric alignment under dynamic poses.
  \item We design a part-based pose embedding mechanism based on local joint rotations and integrate learnable appearance priors to guide the rendering network in accurately completing occluded regions and fine-grained textures.
\end{itemize}

\section{Related work}\label{sec2}

\subsection{3D Human Modeling}

3D human model reconstruction is a crucial topic in computer vision, especially for virtual character synthesis and free-viewpoint video applications, where accurate 3D models are essential. Traditional 3D modeling methods mainly rely on multi-view stereo techniques~\cite{guo2019relightables, schonberger2016structure, schonberger2016pixelwise} or depth fusion~\cite{collet2015high, dou2016fusion4d, su2020robustfusion}, which reconstruct the human geometry from multi-view images or depth sensors. However, these approaches require sophisticated hardware, such as dense camera arrays or depth sensors, which increases the cost and complexity of deployment. To reduce dependence on complex hardware, recent studies have utilized deep learning networks trained on large-scale 3D human datasets to learn prior knowledge, enabling geometry and texture inference from a single image~\cite{saito2019pifu, saito2020pifuhd}. These methods significantly reduce hardware demands, but their performance often degrades under complex poses due to limited diversity in training data. To improve reconstruction accuracy and adaptability, recent methods~\cite{chen2021snarf, mihajlovic2022coap, saito2021scanimate} adopt implicit neural representations to model dynamic humans, enhancing precision through neural optimization. Another line of research combines dynamic implicit neural representations with differentiable rendering to reconstruct 3D human models from video~\cite{kwon2021neural, liu2021neural, remelli2022drivable}. These approaches integrate neural radiance fields and linear blend skinning models, achieving more accurate motion and pose modeling, even under noisy pose inputs~\cite{kwon2021neural, liu2021neural}. Despite their progress, these implicit-based methods typically require long training times and large amounts of high-quality training data, which are often acquired through expensive 3D scanning equipment, increasing data acquisition costs and complexity. Therefore, reducing dependence on high-quality 3D data while maintaining reconstruction fidelity remains a key challenge. In contrast, our method leverages standardized modeling to accelerate the optimization process, enabling high-quality 3D human generation at lower cost.

\subsection{Neural Scene Representations and Rendering}

In recent years, neural scene representations have made significant progress in 3D modeling and rendering, particularly in view synthesis and geometric reconstruction. Neural Radiance Fields (NeRF)~\cite{mildenhall2021nerf} represent a scene as a continuous 5D function and have achieved impressive rendering quality. However, a major limitation of NeRF is the expensive per-scene optimization required, which restricts its scalability and applicability to dynamic scenes. To address this, researchers have proposed generalizable NeRF models~\cite{chen2021mvsnerf, johari2022geonerf, wang2021ibrnet} that learn scene priors conditioned on pixel-aligned features. For instance, MVSNeRF~\cite{chen2021mvsnerf} constructs neural feature volumes using plane-sweep cost volumes and regresses volume density and color via MLPs. Moreover, implicit neural representation methods~\cite{chen2019learning, liu2020dist, park2019deepsdf, sitzmann2019scene, zhi2021place} have been widely used, predicting point-wise scene properties such as occupancy, signed distance, and semantics using MLPs to efficiently represent high-resolution 3D scenes.When combined with NeRF, these approaches enable high-quality image synthesis and have driven advances in rendering quality, speed, and scalability across complex scenes.

\subsection{Neural Radiance Field Rendering from Monocular Video}

High-quality 3D human reconstruction is a core challenge in achieving free-viewpoint video synthesis. Traditional approaches typically rely on multi-view stereo modeling~\cite{guo2019relightables} or depth fusion techniques~\cite{collet2015high}, which require sophisticated equipment for precise capture, making them difficult to deploy in practical applications. To reduce acquisition costs, some methods~\cite{alldieck2022photorealistic,saito2019pifu} attempt to infer human geometry from a single image using neural networks. However, due to limited capability in modeling complex poses and viewpoint variations, their generation quality is constrained in diverse scenarios. 

In recent years, Neural Radiance Fields (NeRF)~\cite{mildenhall2021nerf} have been introduced into human modeling and progressively extended to dynamic scenes. Representative work such as SHERF~\cite{hu2023sherf} estimates a generalizable and animatable human NeRF representation from a single image, significantly simplifying the training process. Nevertheless, it still heavily depends on SMPL~\cite{loper2023smpl} parameters and precise pose estimation, which remain costly to acquire in real-world scenarios. Another line of research~\cite{jiang2023instantavatar,xu2021h,jiang2022neuman,peng2023implicit} focuses on generating human NeRF representations under a canonical T-pose, followed by skeletal or SMPL-driven pose transformations. Among these, A-NeRF~\cite{su2021nerf} initializes the human model using a pre-trained pose estimator, while HumanNeRF~\cite{weng2022humannerf} improves rendering quality through joint optimization of pose and shape representations. However, these methods show limited performance in handling complex motions and occluded scenarios and generally suffer from long training times and heavy reliance on external pose accuracy.Additionally, some research~\cite{hu2022hvtr,jiang2022selfrecon,kwon2021neural} leverages implicit neural representations and differentiable renderers to reconstruct dynamic human representations from videos. Neural Actor~\cite{liu2021neural} incorporates linear blend skinning (LBS)~\cite{lewis2023pose} combined with residual deformation fields to enhance dynamic motion modeling. To mitigate the impact of pose estimation errors, several methods~\cite{su2021nerf,weng2022humannerf} jointly optimize pose and shape representations during training, at the cost of significantly increased training complexity. Recently, part-based modeling strategies have been introduced to accelerate training convergence. For example, NASA~\cite{deng2020nasa} and Coap~\cite{mihajlovic2022coap} adopt part-based implicit functions to model human shapes, while Instant-NVR~\cite{geng2023learning} employs voxelized body-part representations to expedite the optimization process. However, these approaches still face limitations in geometric detail representation and viewpoint generalization. In contrast, our proposed \textbf{MonoPartNeRF} imposes bidirectional deformation constraints on each semantic part and integrates part-level pose and appearance feature fusion, effectively improving geometric reconstruction accuracy and viewpoint consistency. It better adapts to complex pose variations and occluded scenarios, enabling high-quality synthesis of novel views and poses. This provides a more stable and detail-rich solution for free-viewpoint and pose rendering of humans under monocular video settings.

\begin{figure*}[t]
	\centering
	\includegraphics[width=\textwidth]{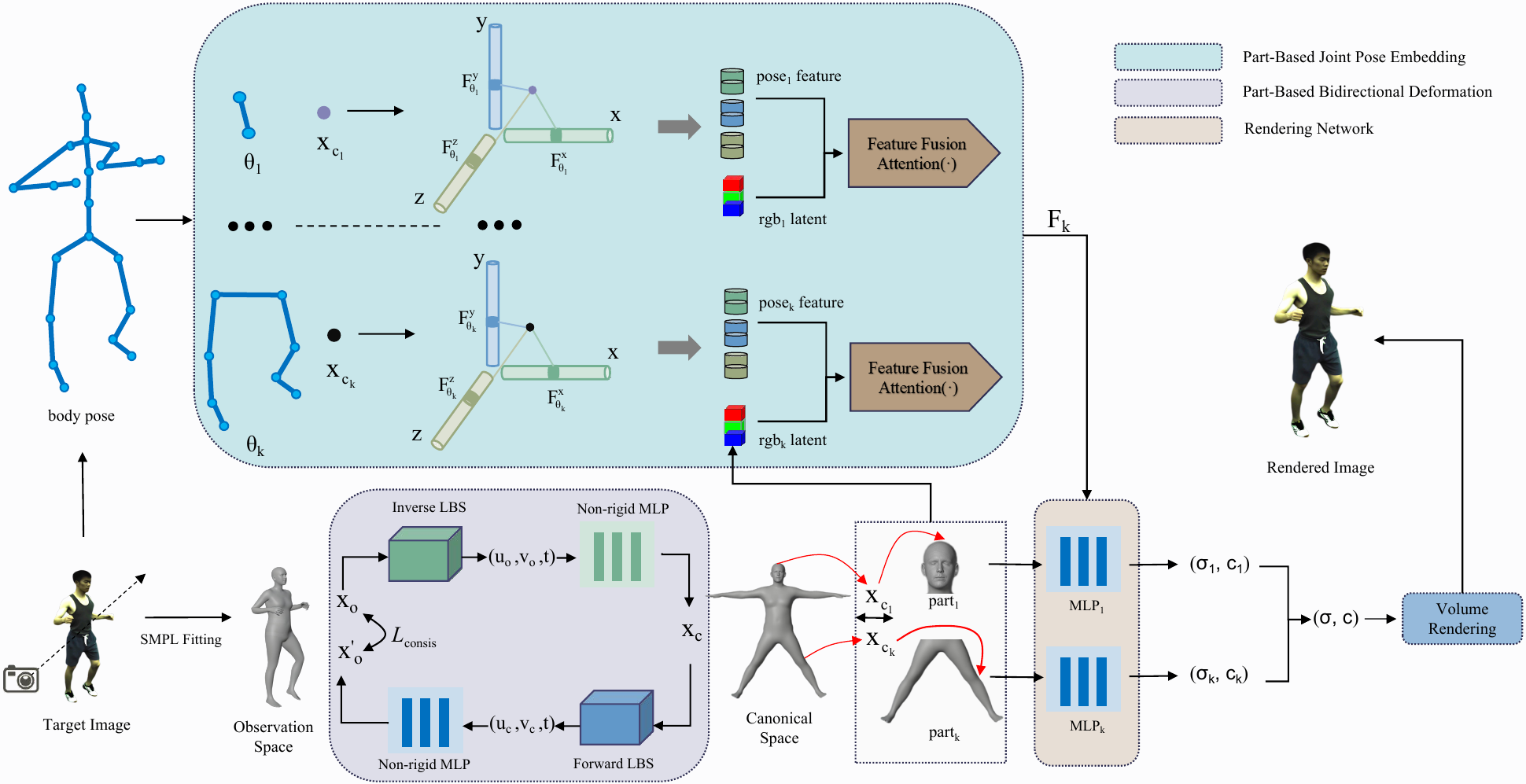}
	\caption{The overall architecture of our method. Given a target image, we first obtain the human pose through SMPL fitting and establish correspondences with the observation space. The human body is segmented into multiple semantic parts, each undergoing bidirectional deformation between the observation space and the canonical space by combining rigid linear blend skinning (LBS) and non-rigid MLP-based deformation, with a consistency loss to ensure accurate alignment across different poses. During the rendering stage, we decompose the global pose vector into local joint rotations, perform similarity search and interpolation for each queried joint to obtain pose embeddings, and fuse regional features using an attention mechanism. Finally, the processed features are fed into part-specific radiance field networks, enabling accurate and coherent image synthesis of novel views and poses under complex deformations and occlusions.}
	\label{fig:2}
\end{figure*}
\section{Method}\label{sec3}
This study aims to address the challenges of unnatural joint connections and geometry loss or texture errors caused by self-occlusion when rendering the human body as multiple parts from a monocular viewpoint. To this end, we propose a novel framework, \textbf{MonoPartNeRF}, as illustrated in Figure~\ref{fig:2}. Specifically, Section~\ref{met:A} describes how the skeletal topology and vertex skinning weights of the SMPL model are used to perform structured semantic partitioning of the human mesh. Section~\ref{met:B} introduces a part-based bidirectional deformation method that incorporates 2D parameterized motion and temporal information, enforcing consistent deformation constraints across different body parts to achieve natural and coherent geometric modeling. Section~\ref{met:C} describes the decomposition of the global pose vector into local joint rotations to obtain part-level pose features, which are then fused with appearance features to guide the rendering network in generating high-fidelity, pose-aware appearances. Section~\ref{met:D} further explains the training objectives and the implementation of volumetric rendering, while Section~\ref{met:E} provides detailed information on the design of the loss functions.

\begin{figure}[t]
	\centering
	\includegraphics[width=0.8\linewidth]{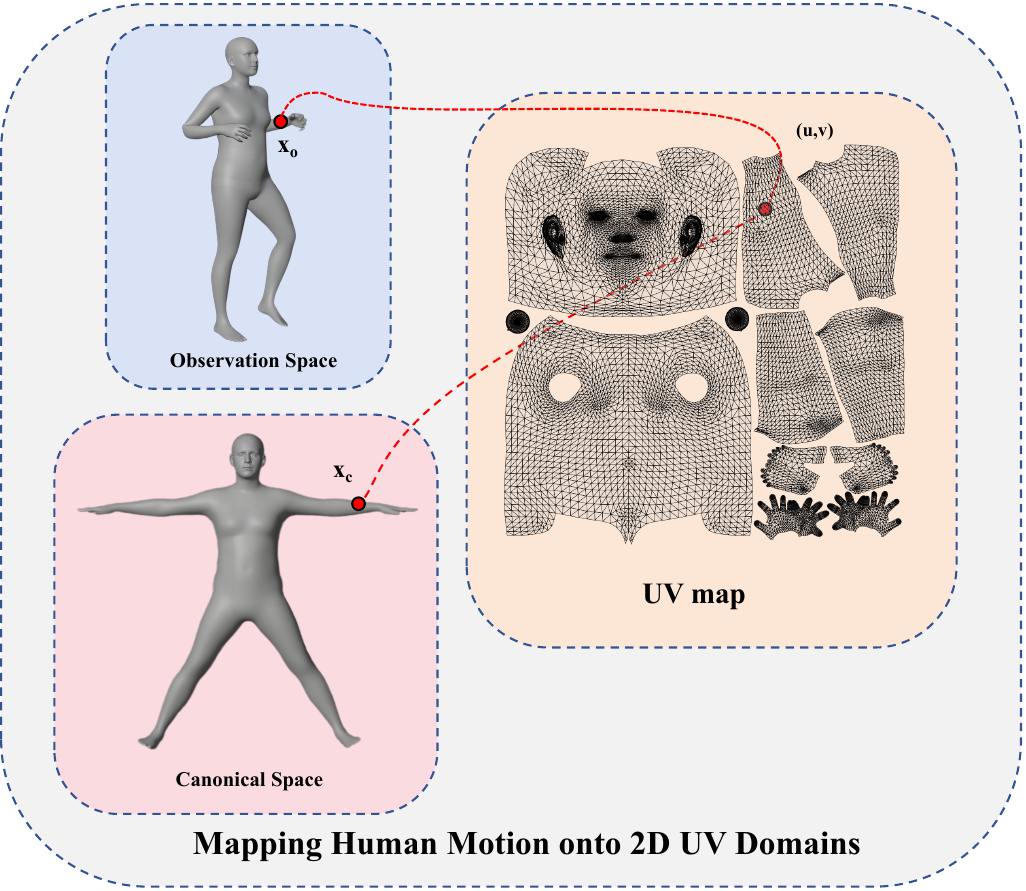}
	\caption{Illustration of motion parameter mapping across multiple spaces. A 3D point \( x_o \) in the observation space and its corresponding point \( x_c \) in the canonical space are jointly mapped to 2D coordinates \( (u, v) \) on the UV map, enabling a unified surface-based motion representation.} 
	\label{fig:3}
\end{figure}
\subsection{Preliminaries}
\label{met:A}

We adopt the SMPL~\cite{loper2023smpl} model as a parametric human prior representation. SMPL represents the human body as a skinned triangular mesh driven by pose and shape parameters, with the surface consisting of $N = 6890$ vertices and the internal skeleton comprising $J = 24$ joints. The deformation of the SMPL mesh is achieved through linear blend skinning (LBS)~\cite{lewis2023pose}, where each vertex $\mathbf{v}_i$ is associated with a fixed set of blend weights $\mathbf{w}_i \in \mathbb{R}^J$ that describe the influence of each joint on that vertex. The mesh is then continuously deformed across frames based on pose parameters. To enhance the modeling capability of both human appearance and motion, we follow the approach of Instant-NVR~\cite{geng2023learning} and introduce a structured semantic partitioning on the SMPL surface. Specifically, based on the skeletal topology of SMPL, each vertex is assigned to the dominant joint corresponding to the maximum value in its blend weights. This results in a decomposition of the entire human mesh into five anatomically meaningful parts: torso, legs, head, left arm, and right arm. This partitioning assigns each vertex a unique semantic label, yielding a structure-aware segmented mesh.

\subsection{Part-Based Bidirectional Deformation}
\label{met:B}

Decomposing the human body into multiple parts for individual rendering allows flexible allocation of model capacity based on the complexity of each body part. However, due to the complex non-rigid structure of the human body, this often leads to unnatural joint connections and difficulty in maintaining coherent shapes in certain regions. The core issue lies in the deformation process from observation space to canonical space, which involves both rigid and non-rigid transformations. During this deformation, some regions may shift, resulting in discontinuities in the rendering process.To address this, we introduce a bidirectional deformation mechanism that enforces consistent deformation constraints between the observation and canonical spaces, ensuring consistent deformation across all body parts. Non-rigid deformation introduces complex and irregular surface changes. Although sampling points may be aligned in 3D space through rigid transformations, directly computing non-rigid offsets using 3D coordinates can lead to feature misalignment or distortion due to pose variation and surface deformation. This is because spatial relationships in 3D space are inherently complex, and local deformations can cause inconsistent offset directions between neighboring sampling points, impairing offset estimation.

As illustrated in Figure~\ref{fig:3}, to improve the accuracy of non-rigid offset computation, we project the 3D coordinates of sampling points onto a 2D parameterized human surface coordinate system \( (u, v) \). This maps the points onto a more regular and continuous 2D surface, making the relative positions more intuitive and easier to handle. Additionally, to capture temporal dynamics, we concatenate the current time frame \( t \) with the parameterized \( (u, v) \) coordinates to form \( (u, v, t) \) as the input feature for each sampling point.The inverse deformation of a sampling point \( \mathbf{x}_{\text{o}} \) in observation space is defined as:
\begin{equation}
    \mathbf{x}_c = \Phi_{\text{LBS}}^{-1}(\mathbf{x}_o) + \Delta \Phi_{\text{MLP}}^{-1}(\mathbf{u}_o, \mathbf{v}_o, \mathbf{t})
\end{equation}
where \( \Phi_{\text{LBS}}^{-1} \) denotes the rigid inverse transformation based on Linear Blend Skinning (LBS)~\cite{lewis2023pose}, and \( \Delta \Phi_{\text{MLP}}^{-1} \) represents the non-rigid deformation predicted by a multi-layer perceptron (MLP), with input being the parameterized coordinates \((\mathbf{u}_o, \mathbf{v}_o)\) and time frame \(\mathbf{t}\).Similarly, the forward deformation of a sampling point \( \mathbf{x}_{\text{c}} \) in canonical space is:
\begin{equation}
    \mathbf{x}_o^{\textbf{'}} = \Phi_{\text{LBS}}(\mathbf{x}_c) + \Delta \Phi_{\text{MLP}}(\mathbf{u}_c, \mathbf{v}_c, \mathbf{t})
\end{equation}
Here, \( \Phi_{\text{LBS}} \) denotes the rigid forward deformation, and \( \Delta \Phi_{\text{MLP}} \) represents the non-rigid offset computed using the corresponding parameterized coordinates and temporal information.This design effectively reduces the original complex 4D space-time representation \( (x, y, z, t) \) into a 3D surface-time domain \( (u, v, t) \). By doing so, the model can distinguish dynamic features of the same surface point across different time steps, avoiding temporal confusion and accurately capturing deformation trends. For instance, slight movements of the same joint at different time frames can be differentiated through the temporal dimension \( t \), ensuring more reasonable offset estimation per frame. This allows sampling points to maintain smooth and continuous transformations in the \( (u, v) \) space, enabling the model to better capture surface details and compute accurate non-rigid deformations.

To enhance the consistency between part-based voxelized human representation and the deformation fields, we define a bidirectional consistency loss as:
\begin{equation}
\mathcal{L}_{\text{consis}} = \{ d \mid d > \theta \}, \quad
d = \|\mathbf{x}_o - \mathbf{x}_o^{\textbf{'}} \|
\end{equation}
where \( \|\cdot\| \) denotes the \( L_2 \) norm (Euclidean distance). This loss measures the consistency of the deformation field during forward and backward mappings, ensuring that corresponding points before and after deformation align correctly, thereby preventing discontinuities and artifacts due to accumulated deformation errors. Furthermore, the threshold hyperparameter \( \theta \) defines an acceptable error margin. The loss contributes only when the offset \( d \) exceeds \( \theta \), thus suppressing local noise and enhancing the stability and robustness of the deformation field.

\subsection{Part-Based Joint Pose Embedding}
\label{met:C}

Under monocular view conditions, the limited field of view often results in certain body parts being self-occluded. This leads to missing visual information during rendering, causing shape distortion or incorrect texture predictions. To address this issue, we introduce a part-based joint pose embedding mechanism that enables the model to independently model different body parts and their associated joints. This allows more accurate inference of the shape and motion trajectories of occluded regions and helps fill in the missing information.Specifically, we divide the human body into several independent parts, such as the head and neck joints, torso and spine, arms and shoulder-elbow-wrist joints, and legs with hip-knee-ankle joints. Each part’s motion is controlled by its corresponding joints, exhibiting localized rather than globally consistent motion patterns. Therefore, we embed joint poses separately for each part to ensure the model can learn motion and deformation patterns specific to that part. We first extract the rotation parameters for each body part based on the joint rotations within that part:
\begin{equation}
    \Theta_{k} = \{ \Theta_{j} \mid j \in \mathcal{J}_k \}
\end{equation}
where \( \Theta_{j} \) represents the rotation parameters of joint \( j \), and \( \mathcal{J}_k \) denotes the set of joints included in the \( k \)-th body part. We first extract all pose samples from the training set and independently compute local pose features for each body part along the x, y, and z axes to construct stable and generalizable pose embedding representations. During the training phase, the model learns the pose features of each body part along the x, y, and z axes in each pose sample. These features are learnable parameters. During training, the model optimizes these features through backpropagation, allowing it to better adapt to pose variations and body movements, thereby gradually improving the model's accuracy and generalization ability. During the testing phase, we retrieve the most similar reference samples from the training set based on a similarity metric with the current target pose. Specifically, the similarity is computed based on the differences between the rotation parameters $\Theta_k$ of each body part to measure the distance between poses. For each body part, we select the top 5 reference samples most similar to the target pose and compute the weighted average of their local pose features along the x, y, and z axes to obtain the target’s local pose features for each axis. Based on this, we take a query point $\mathbf{x}_{\text{o}}$ in the observation space and map it to the corresponding point $\mathbf{x}_{\text{c}}$ in the canonical space via inverse deformation. Then, $\mathbf{x}_{\text{c}}$ is projected onto the x, y, and z axes, and the corresponding features are extracted from the target pose representation via bilinear interpolation. Finally, the multi-axis features are fused to form the pose feature for the query point $\mathbf{x}_{\text{o}}$:
\begin{equation}
    \mathbf{P}_k = \oplus \left( \mathbf{F}_{\theta_{\text{k}}}^{x}, \mathbf{F}_{\theta_{\text{k}}}^{y}, \mathbf{F}_{\theta_{\text{k}}}^{z} \right)
\end{equation}
where \( \oplus \) denotes feature concatenation, and \( \mathbf{F}_{\theta_{\text{k}}}^{\{x,y,z\}} \) represents features sampled via bilinear interpolation along the x, y, and z axes for the \( k \)-th body part. This axis-based decomposition allows the model to more accurately capture motion variations along each direction and convert them into learnable representations, thereby improving its understanding of pose changes for each part.

To further enhance the expressive power of the pose features, we introduce a set of learnable feature vectors $\mathbf{rgb\_latent}$ to model the appearance of the human body. These feature representations capture appearance attributes that are gradually learned by the model during training and effectively complement the missing texture and color cues in the pose encoding, thereby improving the final rendering quality. When the model encounters occluded or under-observed regions, it can extract the most relevant information from these appearance vectors based on the current pose features as supplementary guidance.Specifically, we fuse \( \mathbf{rgb\_latent} \) with the pose embeddings using a cross-attention mechanism that adaptively associates the motion state of each part with its corresponding prior appearance features:
\begin{equation}
\begin{aligned}
\mathbf{F}_k = \text{Attention}(
&Q = \mathbf{P}_k, \\
&K = \{\mathbf{rgb\_latent}\}, \\
&V = \{\mathbf{rgb\_latent}\}).
\end{aligned}
\end{equation}
This mechanism allows the model to adaptively select the most relevant appearance prior for each part’s current pose state, thereby improving rendering quality and consistency in occluded or under-observed scenarios.

Our part-based joint pose embedding approach enables the model to flexibly allocate computational resources according to the complexity of each part. For example, the head and hands require more fine-grained representation due to their rich detail and complex motion, while relatively stable regions such as the torso are modeled in a more simplified manner to reduce computational overhead. This strategy ensures high reconstruction accuracy while improving rendering efficiency, resulting in more natural and fluid dynamic human representations.

\subsection{Neural Rendering Process}
\label{met:D}

We perform part-based neural rendering of the human body in the canonical space. For each body part \( k \), we use an independent rendering network \( \text{MLP}_k \) to predict the volume density \( \sigma_k \) and color \( c_k \) at a query point \( \mathbf{x}_{c_k} \). The input to each network includes the 3D position and the fused feature vector \( \mathbf{F}_k \), which is obtained by combining part-specific pose features and time-dependent appearance features via a cross-attention mechanism. The rendering function is defined as:
\begin{equation}
(\sigma_k, c_k) = \text{MLP}_k(\psi_k(\mathbf{x}_{c_k}), \mathbf{F}_k)
\end{equation}
where $\psi_k$ denotes the multi-resolution hash encoding applied to the $k$-th body part, used to efficiently represent the input 3D spatial coordinate $\mathbf{x}_{c_k}$. This encoding strategy significantly reduces computational cost while preserving expressive power, thereby accelerating network convergence during training. Considering that the query point $\mathbf{x}_{c_k}$ may lie near the boundaries of multiple body parts, we feed it into all $K$ rendering networks $\text{MLP}_k$ corresponding to each body part to obtain multiple predictions ${(\sigma_k, c_k)}_{k=1}^K$. Among these candidate results, we select the one with the highest density $\sigma_k$ as the final output for the point:
\begin{equation}
\sigma = \max_k \sigma_k, \quad c = c_{\arg\max_k \sigma_k}
\end{equation}
The core rationale behind this selection strategy is that the volume density \( \sigma_k \) reflects the likelihood of the query point belonging to the geometric region of body part \( k \). A higher density indicates a stronger association with that part. By selecting the result with the maximum density, we can avoid conflicting predictions and ensure that the rendering is based on the most representative network output, thereby enhancing continuity and physical consistency in boundary regions.

After obtaining the final density \( \sigma \) and color \( c \) for all sampled points, we apply the volume rendering method~\cite{mildenhall2021nerf} to composite the pixel color along the ray \( r \) as follows:
\begin{equation}
\begin{aligned}
C(r) &= \sum_{i=1}^D T_i \cdot \alpha_i \cdot c(\mathbf{x}_i), \\
\alpha_i &= 1 - \exp(-\sigma(\mathbf{x}_i) \cdot \Delta t_i), \\
T_i &= \prod_{j=1}^{i-1}(1 - \alpha_j)
\end{aligned}
\end{equation}
where \( \alpha_i \) denotes the opacity of the \( i \)-th sampled point, \( \Delta t_i \) is the interval between adjacent samples, and \( T_i \) is the accumulated transmittance from the ray origin to the \( i \)-th sample. This volume rendering process aggregates predictions from all body parts into a complete human image, ensuring structural continuity and appearance consistency across parts, thereby achieving high-fidelity dynamic human rendering.

\subsection{Loss Design}
\label{met:E}

To optimize the performance of our model, we design a composite loss function based on the discrepancy between the rendered and ground-truth images. It includes reconstruction loss, perceptual loss, and consistency loss. The overall objective is defined as:
\begin{equation}
\mathcal{L}_{\text{total}} = \lambda_{\text{m}} \mathcal{L}_{\text{MSE}} + \lambda_{\text{p}} \mathcal{L}_{\text{LPIPS}} + \lambda_{\text{c}} \mathcal{L}_{\text{CONSIS}}
\end{equation}
where \( \mathcal{L}_{\text{MSE}} \) is the pixel-wise mean squared error that enforces accurate color reconstruction;
\( \mathcal{L}_{\text{LPIPS}} \)~\cite{johnson2016perceptual} is the perceptual loss that measures similarity in a high-level semantic feature space;
and \( \mathcal{L}_{\text{CONSIS}} \) is the consistency loss, which encourages geometric continuity and appearance smoothness at the boundaries between different body parts. This loss effectively prevents artifacts such as discontinuities, seams, or color jumps. The weighting coefficients for the above loss terms are set as $\lambda_{\text{m}}=1$, $\lambda_{\text{p}}=1$, and $\lambda_{\text{c}}=1$. The proposed optimization strategy simultaneously improves reconstruction accuracy, perceptual quality, and structural consistency of the human body.

\section{Experiment}

\subsection{Datasets}

Following the experimental setup of Instant-NVR~\cite{geng2023learning}, we conduct training and evaluation on two datasets: ZJU-MoCap~\cite{geng2023learning,peng2023implicit} and MonoCap~\cite{peng2024animatable}.

For the ZJU-MoCap dataset, we select 6 subjects (377, 386, 387, 392, 393, 394). During training, we use images captured from only a single camera as input, while in the evaluation phase, we adopt images from different viewpoints to verify the model's generalization capability across viewpoints. Each subject is sampled at a rate of 1 frame every 5 frames, with 100 frames selected for training. To further assess the model's generalization to unseen poses, we partition the data between the training set and the novel-pose evaluation set in approximately a 2:1 frame ratio. This dataset provides high-quality multi-view synchronized images and precise 3D human geometry, offering an ideal platform to evaluate the robustness and generalization of our method in multi-view scenarios.

For the MonoCap dataset, we used the complete data from 4 subjects (lan, marc, olek, vlad), strictly following the camera setup described in \cite{geng2023learning}. The MonoCap dataset consists of two subsets: DynaCap \cite{habermann2021real} and DeepCap \cite{habermann2020deepcap}, which respectively cover dynamic motion capture data and RGB-D clothing deformation data. These subsets feature diverse motion variations, viewpoint changes, and occlusions, broadly capturing complex real-world scenarios and human body deformations, thereby enhancing the diversity and challenge of the dataset. The training data was also sampled every 5 frames, selecting 100 frames for training. A frame ratio of approximately 2:1 was maintained between the training set and the novel-pose evaluation set to systematically assess the model’s generalization ability to unseen poses.

\begin{table*}[t]
\centering
\scriptsize
\renewcommand{\arraystretch}{1.1}
\setlength{\tabcolsep}{6pt}
\begin{tabular}{l|c|ccc|ccc}
\toprule
\multirow{2}{*}{\textbf{Methods}} & \multirow{2}{*}{\textbf{Training Time}} & \multicolumn{3}{c|}{\textbf{ZJU-MoCap}} & \multicolumn{3}{c}{\textbf{MonoCap}} \\
\cline{3-8} \\[-2.5ex]
& & PSNR↑ & SSIM↑ & LPIPS*$\downarrow$ & PSNR↑ & SSIM↑ & LPIPS*$\downarrow$ \\
\midrule
NeuralBody \cite{peng2021neural}    & 10 h & 29.03 & 0.964 & 42.47 & 32.36 & 0.986 & 16.70 \\
SHERF \cite{hu2023sherf}            & 10 h & 29.96 & 0.968 & 40.63 & 32.41 & 0.986 & 16.41 \\
HumanNeRF \cite{weng2022humannerf}  & 10 h & 30.66 & 0.969 & 33.38 & 32.68 & 0.987 & 15.52 \\
Instant-NVR \cite{geng2023learning} & 5 min & 31.01 & 0.971 & 38.45 & 32.61 & \textbf{0.988} & 16.68 \\
GauHuman \cite{hu2024gauhuman}      & 2 min & 31.34 & 0.965 & \textbf{30.51} & 33.45 & 0.985 & \textbf{13.15} \\
\midrule
\textbf{Ours}                       & 6 min & \textbf{32.12} & \textbf{0.972} & 35.50 & \textbf{33.67} & \textbf{0.988} & 14.87 \\
\bottomrule
\end{tabular}
\caption{
Quantitative comparison results on novel views across the ZJU-MoCap and MonoCap datasets, with all metrics reported as the average across different subjects in each dataset. \textbf{Note:} For clarity of presentation, the LPIPS* metric has been scaled by a factor of $10^3$.
}
\label{tab:1}
\end{table*}

\begin{table*}[t]
\centering
\scriptsize
\renewcommand{\arraystretch}{1.1}
\setlength{\tabcolsep}{6pt}  
\begin{tabular}{l|c|ccc|ccc}
\toprule
\multirow{2}{*}{\textbf{Methods}} & \multirow{2}{*}{\textbf{Training Time}} & \multicolumn{3}{c|}{\textbf{ZJU-MoCap}} & \multicolumn{3}{c}{\textbf{MonoCap}} \\
\cline{3-8} \\[-2.5ex]
& & PSNR↑ & SSIM↑ & LPIPS*$\downarrow$ & PSNR↑ & SSIM↑ & LPIPS*$\downarrow$ \\
\midrule
NeuralBody \cite{peng2021neural}    & 10 h & 29.17 & 0.966 & 42.21 & 32.62 & 0.986 & 16.44 \\
SHERF \cite{hu2023sherf}            & 10 h & 29.75 & 0.968 & 39.12 & 32.86 & 0.987 & 15.98 \\
HumanNeRF \cite{weng2022humannerf}  & 10 h & 30.88 & 0.974 & 31.21 & 33.19 & 0.988 & 13.29 \\
Instant-NVR \cite{geng2023learning} & 5 min & 31.53 & 0.976 & 31.60 & 33.58 & \textbf{0.989} & 13.14 \\
GauHuman \cite{hu2024gauhuman}      & 2 min & 31.66 & 0.969 & \textbf{28.94} & 33.84 & 0.987 & 12.17 \\
\midrule
\textbf{Ours}                       & 6 min & \textbf{32.42} & \textbf{0.977} & 29.05 & \textbf{33.91} & \textbf{0.989} & \textbf{12.03} \\
\bottomrule
\end{tabular}
\caption{
Quantitative comparison results on novel poses across the ZJU-MoCap and MonoCap datasets, with all metrics reported as the average across different subjects in each dataset. \textbf{Note:} For clarity of presentation, the LPIPS* metric has been scaled by a factor of $10^3$.
}
\label{tab:2}
\end{table*}

\subsection{Implementation Details}
All models were implemented using the PyTorch framework~\cite{paszke2019pytorch} and trained on a single NVIDIA RTX 3090 GPU. We used a pre-fitted SMPL model~\cite{loper2023smpl} as a human geometry prior, obtained via existing methods. This SMPL prior remained fixed during training to provide an initial volumetric guide. We adopted the Adam optimizer~\cite{kingma2014adam} with a learning rate of $5\times10^{-4}$. The input image resolution was set to $512 \times 512$. The network takes SMPL parameters as conditional input and fits multi-view, multi-pose human images through volumetric rendering, thereby enabling efficient reconstruction of dynamic human bodies.

\begin{figure*}[t]
	\centering
	\includegraphics[width=\textwidth]{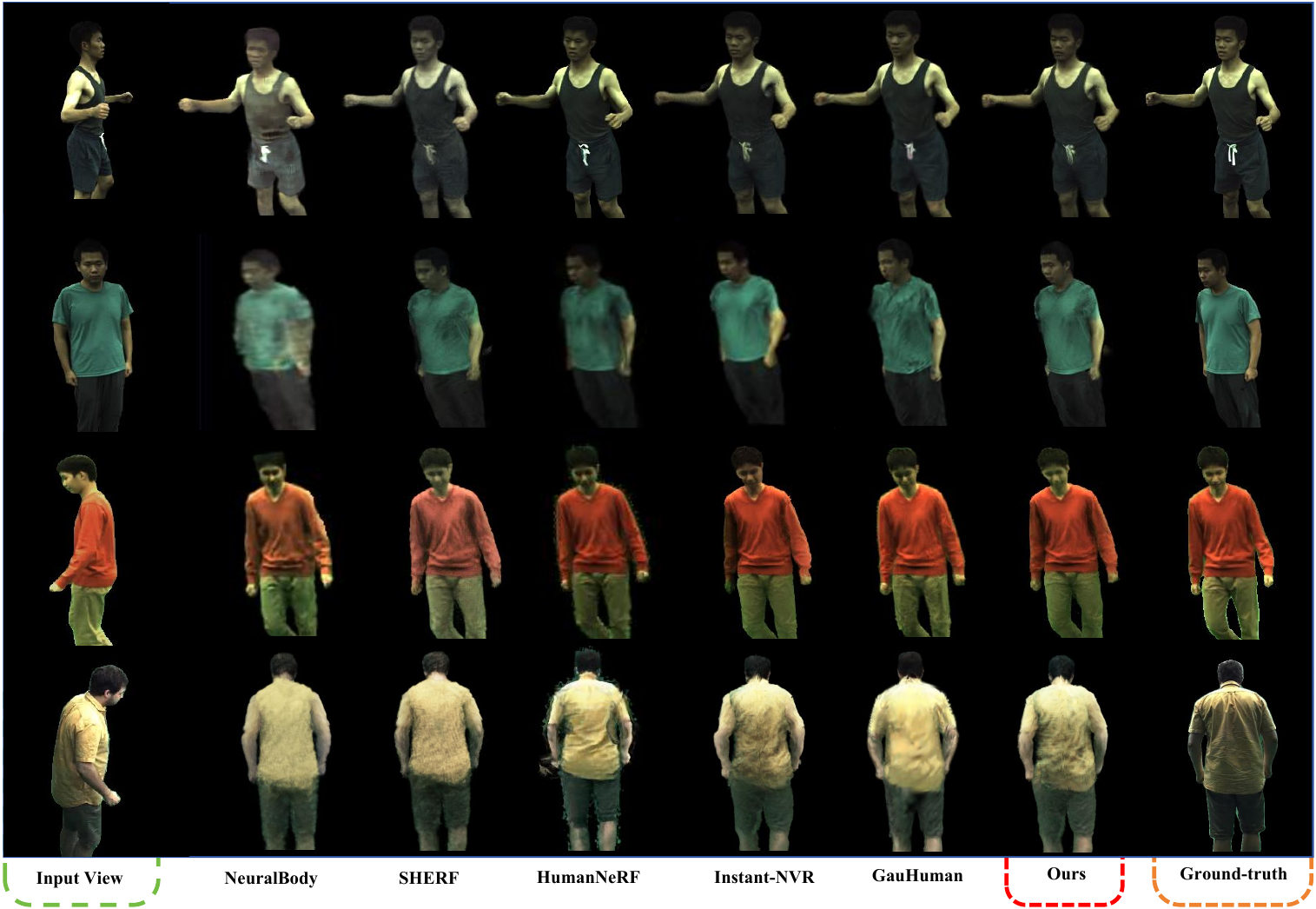}
	\caption{Qualitative results of novel view synthesis on the ZJU-MoCap and MonoCap datasets.}
	\label{fig:4}
\end{figure*}

\begin{figure*}[t]
	\centering
	\includegraphics[width=\textwidth]{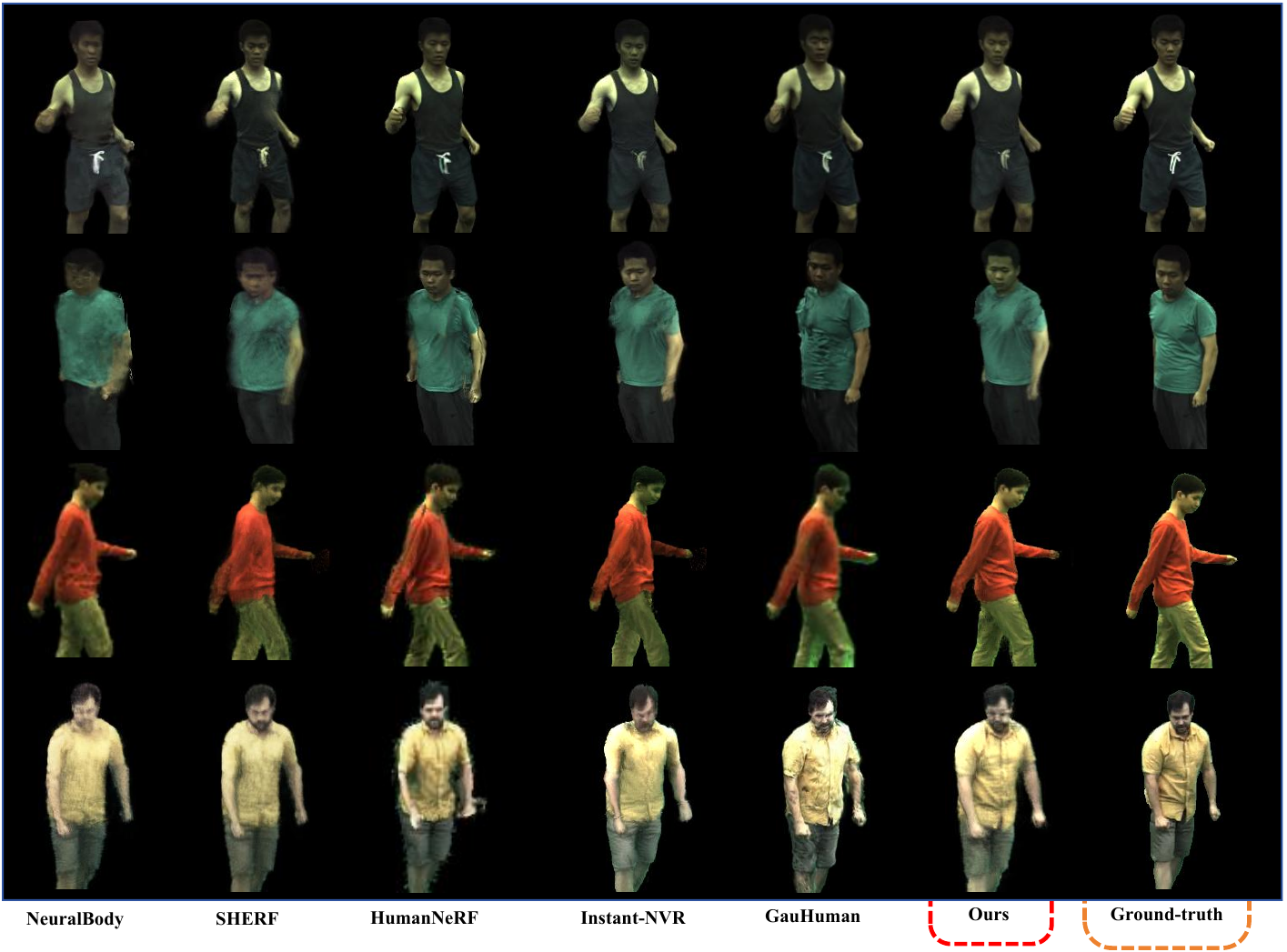}
	\caption{Qualitative results of novel pose synthesis on the ZJU-MoCap and MonoCap datasets.}
	\label{fig:5}
\end{figure*}

\begin{figure}[t]
	\centering
	\includegraphics[width=\textwidth]{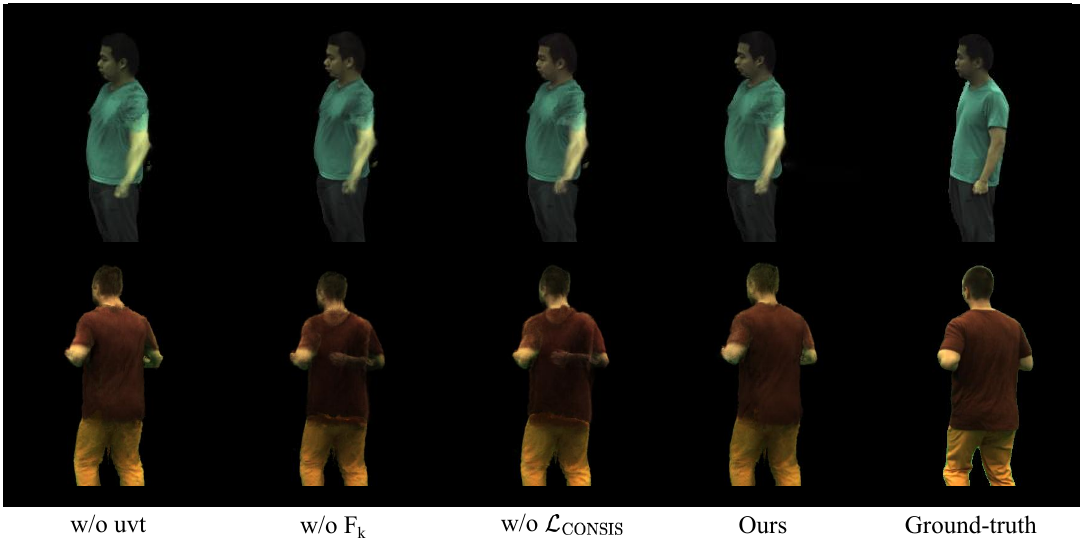}
	\caption{Qualitative results of the ablation study on novel view synthesis on the ZJU-MoCap and MonoCap datasets.}
	\label{fig:6}
\end{figure}

\begin{figure}[t]
	\centering
	\includegraphics[width=\textwidth]{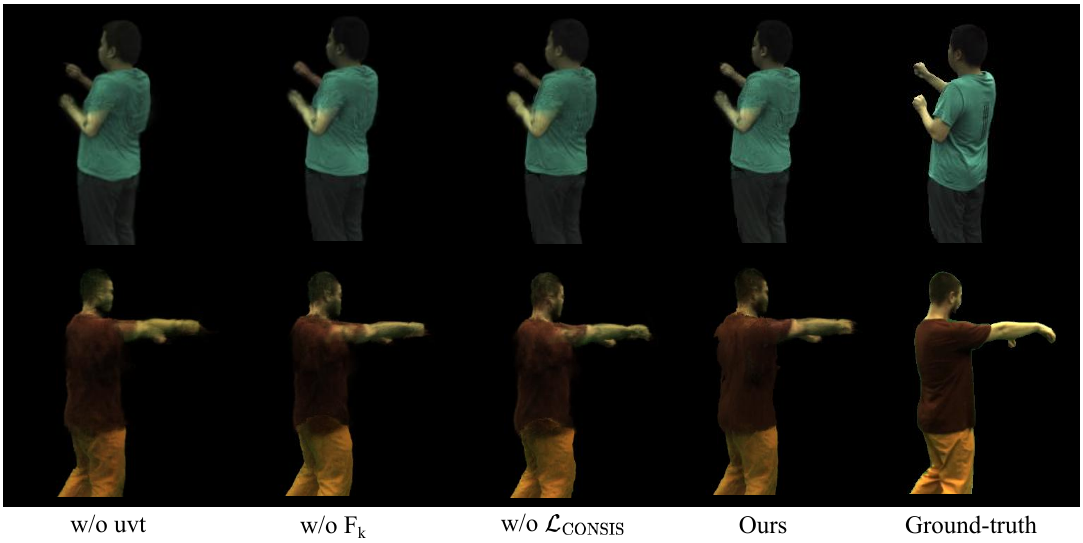}
	\caption{Qualitative results of the ablation study on novel pose synthesis on the ZJU-MoCap and MonoCap datasets.}
	\label{fig:7}
\end{figure}

\begin{figure}[t]
	\centering
	\includegraphics[width=\textwidth]{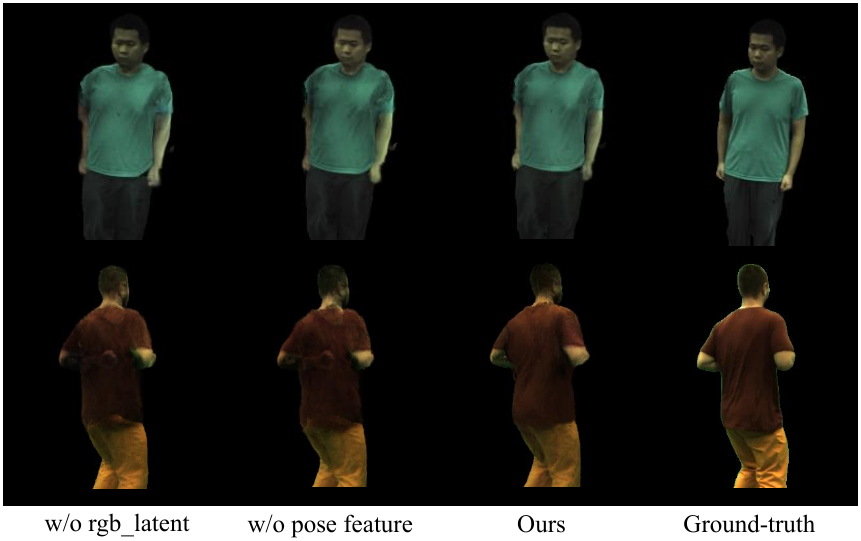}
	\caption{Qualitative ablation results on the ZJU-MoCap and MonoCap datasets evaluating the independent effects of pose and appearance features in novel view synthesis.}
	\label{fig:8}
\end{figure}

\subsection{Comparison with State-of-the-Art Methods}

\subsubsection{Baselines}
We selected a range of representative state-of-the-art methods as comparative baselines, covering the two major directions in dynamic human modeling: novel view synthesis and novel pose synthesis. These methods reflect the current advancements in modeling strategies, input settings, and geometric representations for dynamic human reconstruction.

Specifically, GauHuman~\cite{hu2024gauhuman} adopts a hybrid strategy that combines Gaussian volumetric representation with pose-conditioned modeling, demonstrating superior performance in both dynamic pose accuracy and rendering efficiency. Instant-NVR~\cite{geng2023learning} employs a region-based modeling approach, enhancing adaptability to structural variations in the human body through multi-semantic partitioning, and achieves competitive reconstruction accuracy and training efficiency. HumanNeRF~\cite{weng2022humannerf} builds dynamic human models based on conditional Neural Radiance Fields, primarily designed for multi-view inputs. It offers certain generalization capability across poses but has limitations under monocular or highly articulated pose scenarios. SHERF~\cite{hu2023sherf} estimates a controllable human NeRF representation from a single image, enabling pose manipulation and free-view rendering, making it suitable for data-scarce environments. NeuralBody~\cite{peng2021neural} leverages an SMPL-driven implicit volumetric representation, effectively reconstructing dynamic humans under multi-view supervision, though it shows limited adaptability to extreme pose variations and large viewpoint changes.

\subsubsection{Quantitative Results}
We conducted quantitative experiments on novel view synthesis and novel pose synthesis tasks using the ZJU-MoCap and MonoCap datasets. The results are summarized in Table~\ref{tab:1} and Table~\ref{tab:2}. All evaluation metrics are averaged across different test sequences. Note that the LPIPS* metric has been scaled by $10^3$ for better readability. Although our method is not always the best on every individual sequence, it consistently performs competitively across PSNR, SSIM, and LPIPS metrics, achieving or approaching state-of-the-art results. A higher PSNR indicates lower pixel-wise reconstruction error, SSIM reflects improved structural consistency, and a lower LPIPS value suggests greater perceptual similarity to real images. Together, these results demonstrate the superiority of our method in terms of image fidelity, structural accuracy, and visual realism.

In addition, our method also demonstrates superior training efficiency. Compared to methods such as NeuralBody~\cite{peng2021neural}, SHERF~\cite{hu2023sherf}, and HumanNeRF~\cite{weng2022humannerf}, which require approximately 10 hours of training, and Instant-NVR~\cite{geng2023learning}, which takes 5 minutes, our approach completes training in just 6 minutes, significantly reducing training costs while maintaining strong performance. Although GauHuman~\cite{hu2024gauhuman} achieves faster training speed (2 minutes), its performance in terms of image quality and structural continuity still falls short of ours. Overall, our method strikes a better balance between accuracy and efficiency, offering strong practicality and scalability.

Compared to existing baselines, our approach shows clear advantages. NeuralBody~\cite{peng2021neural}, which utilizes an SMPL-driven implicit volumetric representation, struggles under monocular input when dealing with complex poses and occluded regions, often resulting in blurry details and unstable pose transfer. SHERF~\cite{hu2023sherf} can generate controllable human representations from a single image and shows progress in occlusion modeling, but it lacks robustness in maintaining geometric continuity under large pose variations, leading to visible discontinuities and artifacts. HumanNeRF~\cite{weng2022humannerf}, based on conditional neural radiance fields, performs reasonably well under multi-view settings but lacks strong geometric priors, resulting in local distortions when applied to monocular inputs.

Instant-NVR~\cite{geng2023learning} and GauHuman~\cite{hu2024gauhuman} represent the latest advancements in this domain. The former adopts a semantic part-based modeling strategy to improve efficiency but often suffers from unnatural joint connections and local shape distortions under pose variations, especially in monocular settings. The latter combines Gaussian volumetric representation with pose-conditioned modeling and shows strong performance in motion representation. However, it still falls short in recovering occluded regions and texture details, leading to noticeable gaps in geometric precision and visual quality compared to our method.

In contrast, our method is built on a part-based rendering paradigm that allocates parameter density according to the structural complexity of each body part, thereby enhancing the modeling of local details. Furthermore, our bidirectional deformation mechanism and consistency loss ensure accurate alignment of body parts under varying poses into a unified canonical space, achieving pose-geometry disentanglement and preserving natural and continuous joint connections. To address the challenge of occlusion reconstruction under monocular views, we design a part-aware pose embedding mechanism that decomposes the global pose representation into localized joint rotations, and leverages similar-pose interpolation to achieve fine-grained feature alignment. Finally, we fuse 3D pose features with local appearance codes to guide neural rendering, producing realistic and detail-rich images, and achieving superior performance on both novel view and novel pose synthesis tasks.

\subsubsection{Qualitative Results}
Figures~\ref{fig:4} and~\ref{fig:5} present qualitative comparisons between our method and existing approaches on the ZJU-MoCap and MonoCap datasets, corresponding to the novel view and novel pose settings, respectively. Compared to other methods, our approach demonstrates superior visual quality and consistency in terms of structural preservation, boundary sharpness, and texture details.

In the novel view synthesis task shown in Figure~\ref{fig:4}, GauHuman~\cite{hu2024gauhuman} exhibits certain advantages in overall structural reconstruction but still suffers from blurry details and localized artifacts, especially in occluded regions. Instant-NVR~\cite{geng2023learning} shows decent semantic modeling capability but often produces unnatural joint connections and structural discontinuities. HumanNeRF~\cite{weng2022humannerf} and SHERF~\cite{hu2023sherf} suffer from varying degrees of texture distortion and detail loss, particularly in occluded body parts. NeuralBody~\cite{peng2021neural} generally yields lower rendering quality with blurry edges and noticeable color shifts. In contrast, our method achieves more natural results in terms of structural alignment, boundary preservation, and local texture recovery. It effectively avoids color drifting and geometric distortions, maintaining stable and clear performance even in occluded and complex regions, resulting in coherent and high-quality reconstructions.

In the novel pose synthesis task shown in Figure~\ref{fig:5}, our method also demonstrates superior generalization and structural consistency. Although other methods can produce reasonable results within the range of training poses, they struggle under unseen complex poses: GauHuman suffers from local deformations and missing occlusion details, Instant-NVR fails to maintain natural joint connections, and HumanNeRF, SHERF, and NeuralBody often produce pose distortions, texture blurring, or edge breakages. In comparison, our approach maintains accurate joint alignment, consistent textures, and sharp boundaries even under unseen poses, producing more natural and continuous results, further validating its robustness and generalization to challenging pose variations.

In summary, across both novel view and novel pose synthesis tasks, our method significantly outperforms existing approaches in multiple aspects, showcasing stronger structural preservation, visual fidelity, and detail recovery.

\begin{table}[h]
\centering
\caption{Quantitative results of the ablation study on novel view synthesis across the ZJU-MoCap and MonoCap datasets. All metrics are reported as the average across different subjects in each dataset.}
\label{tab:3}
\resizebox{\linewidth}{!}{
\begin{tabular}{l|ccc|ccc}
\toprule
 & \multicolumn{3}{c|}{\textbf{ZJU-MoCap}} & \multicolumn{3}{c}{\textbf{MonoCap}} \\
\addlinespace[2pt]
\midrule
\textbf{Methods} & \textbf{PSNR$\uparrow$} & \textbf{SSIM$\uparrow$} & \textbf{LPIPS*$\downarrow$} & \textbf{PSNR$\uparrow$} & \textbf{SSIM$\uparrow$} & \textbf{LPIPS*$\downarrow$} \\
\addlinespace[2pt]
\midrule
w/o uvt & 32.03 & 0.972 & 35.78 & 33.57 & 0.988 & 14.96 \\
w/o $\text{F}_k$ & 31.79 & 0.972 & 36.14 & 33.52 & 0.988 & 15.03 \\
w/o $\mathcal{L}_{\text{CONSS}}$ & 31.63 & 0.971 & 36.75 & 33.12 & 0.988 & 15.77 \\
\addlinespace[2pt]
\midrule
\textbf{ours} & \textbf{32.12} & \textbf{0.972} & \textbf{35.50} & \textbf{33.67} & \textbf{0.988} & \textbf{14.87} \\
\bottomrule
\end{tabular}
}
\end{table}

\begin{table}[h]
\centering
\caption{Quantitative results of the ablation study on novel pose synthesis across the ZJU-MoCap and MonoCap datasets. All metrics are reported as the average across different subjects in each dataset.}
\label{tab:4}
\resizebox{\linewidth}{!}{
\begin{tabular}{l|ccc|ccc}
\toprule
 & \multicolumn{3}{c|}{\textbf{ZJU-MoCap}} & \multicolumn{3}{c}{\textbf{MonoCap}} \\
\addlinespace[2pt] 
\midrule
\textbf{Methods} & \textbf{PSNR$\uparrow$} & \textbf{SSIM$\uparrow$} & \textbf{LPIPS*$\downarrow$} & \textbf{PSNR$\uparrow$} & \textbf{SSIM$\uparrow$} & \textbf{LPIPS*$\downarrow$} \\
\addlinespace[2pt] 
\midrule
w/o uvt & 32.26 & 0.977 & 29.41 & 33.83 & 0.989 & 12.33 \\
w/o $\text{F}_k$ & 32.03 & 0.977 & 30.16 & 33.78 & 0.989 & 12.64 \\
w/o $\mathcal{L}_{\text{CONSS}}$ & 31.94 & 0.976 & 30.82 & 33.74 & 0.989 & 12.88 \\
\addlinespace[2pt] 
\midrule
\textbf{ours} & \textbf{32.42} & \textbf{0.977} & \textbf{29.05} & \textbf{33.91} & \textbf{0.989} & \textbf{12.03} \\
\bottomrule
\end{tabular}
}
\end{table}

\begin{table}[h]
\centering
\caption{Quantitative ablation results on the ZJU-MoCap and MonoCap datasets evaluating the independent effects of pose and appearance features in novel view synthesis. All metrics are reported as the average across different subjects in each dataset.}
\label{tab:5}
\resizebox{\linewidth}{!}{
\begin{tabular}{l|ccc|ccc}
\toprule
 & \multicolumn{3}{c|}{\textbf{ZJU-MoCap}} & \multicolumn{3}{c}{\textbf{MonoCap}} \\
\addlinespace[2pt] 
\midrule
\textbf{Methods} & \textbf{PSNR$\uparrow$} & \textbf{SSIM$\uparrow$} & \textbf{LPIPS*$\downarrow$} & \textbf{PSNR$\uparrow$} & \textbf{SSIM$\uparrow$} & \textbf{LPIPS*$\downarrow$} \\
\addlinespace[2pt] 
\midrule
w/o $\mathbf{rgb\_latent}$ & 31.93 & 0.972 & 35.95 & 33.61 & 0.988 & 14.94 \\
w/o pose feature & 31.88 & 0.972 & 36.03 & 33.58 & 0.988 & 14.97 \\
\addlinespace[2pt] 
\midrule
\textbf{ours} & \textbf{32.12} & \textbf{0.972} & \textbf{35.50} & \textbf{33.67} & \textbf{0.988} & \textbf{14.87} \\
\bottomrule
\end{tabular}
}
\end{table}

\begin{table}[h]
\centering
\caption{Ablation study on the ZJU-MoCap dataset analyzing the effects of varying loss weights ($\lambda_{\text{m}}, \lambda_{\text{p}}, \lambda_{\text{c}}$) on novel view and novel pose synthesis performance. All results are reported as averages over different subjects.}
\label{tab:6}
\resizebox{\linewidth}{!}{
\begin{tabular}{ccc|ccc|ccc}
\toprule
\multicolumn{3}{c|}{\textbf{Loss Weights}} & \multicolumn{3}{c|}{\textbf{Novel View}} & \multicolumn{3}{c}{\textbf{Novel Pose}} \\
\midrule
\textbf{$\lambda_{\text{m}}$} & \textbf{$\lambda_{\text{p}}$} & \textbf{$\lambda_{\text{c}}$} & \textbf{PSNR$\uparrow$} & \textbf{SSIM$\uparrow$} & \textbf{LPIPS*$\downarrow$} & \textbf{PSNR$\uparrow$} & \textbf{SSIM$\uparrow$} & \textbf{LPIPS*$\downarrow$} \\
\midrule
1.0 & 1.0 & 0.5 & 31.74 & 0.969 & 37.23 & 31.55 & 0.976 & 30.72 \\
1.0 & 0.5 & 1.0 & 31.86 & 0.970 & 35.96 & 31.97 & 0.976 & 29.88 \\
0.5 & 1.0 & 1.0 & 31.52 & 0.971 & 36.47 & 32.15 & 0.975 & 30.13 \\
\midrule
\textbf{1.0} & \textbf{1.0} & \textbf{1.0} & \textbf{32.12} & \textbf{0.972} & \textbf{35.50} & \textbf{32.42} & \textbf{0.977} & \textbf{29.05} \\
\bottomrule
\end{tabular}
}
\end{table}

\subsection{Ablation Study}
To verify the effectiveness of key components in MonoPartNeRF, we conduct ablation studies on the ZJU-MoCap and MonoCap datasets under both novel view and novel pose settings. Through qualitative and quantitative analyses, we comprehensively evaluate the impact of each module on the final rendering performance, as shown in Figures~\ref{fig:6} and~\ref{fig:7}, and Tables~\ref{tab:3} and~\ref{tab:4}.

\textbf{Consistency Loss.} This loss enforces alignment between deformation processes of different body parts, maintaining geometric coherence across poses and ensuring natural joint transitions. Removing this term leads to structural discontinuities and unnatural transitions during motion.

\textbf{Pose-Appearance Feature Fusion.} We introduce an attention-based mechanism to encode decomposed joint rotations into pose features, which are then jointly modeled with appearance features to produce pose-aware fused representations. These fused features guide the rendering network to generate more detailed and structurally consistent results. This fusion mechanism effectively enhances the model’s ability to reconstruct under complex pose variations and occlusions, especially in occluded regions where it demonstrates strong detail completion capabilities. Removing this module leads to the absence of pose prior guidance, which can result in blurred textures, unnatural deformations, or even incorrect reconstructions, thereby validating its critical role in photorealistic rendering.

\textbf{UVT Spatiotemporal Modeling.} Introducing parameterized UV coordinates and temporal information into the bidirectional non-rigid deformation process improves stable surface modeling. Compared to direct XYZ-based modeling, the UVT representation better preserves texture bindings with the body surface, significantly reducing texture misalignment and surface drift, especially under large pose changes.

\textbf{Independent Effects of Pose and Appearance Features.} To further analyze the individual contributions of pose and appearance features in the rendering network, we conduct separate evaluations by removing either the pose feature or the appearance feature (denoted as $\mathbf{rgb\_latent}$), without using the attention-based fusion mechanism. Figure~\ref{fig:8} presents qualitative results of novel view synthesis on the ZJU-MoCap and MonoCap datasets. It can be observed that without pose features, the model struggles to reconstruct fine details in limb structures and occluded regions; while removing appearance features leads to blurry textures and inaccurate colors, both clearly inferior to the full model. Additionally, Table~\ref{tab:5} provides the corresponding quantitative evaluation in terms of PSNR, SSIM, and LPIPS$^*$. The results show that on both datasets, the full model (\textbf{Ours}) consistently outperforms the variants lacking either feature across all metrics, indicating that pose and appearance features are complementary in improving reconstruction quality and perceptual consistency. This experiment demonstrates the importance of both feature types in high-fidelity rendering and further emphasizes the necessity of the subsequent fusion mechanism design.

\textbf{Effect of Loss Weight Combinations.} To evaluate the relative importance of each loss component in the overall optimization objective, we conduct ablation studies with different loss weight configurations on the ZJU-MoCap dataset. Specifically, we vary the weighting coefficients $\lambda_{\text{m}}$, $\lambda_{\text{p}}$, and $\lambda_{\text{c}}$ for the pixel-wise reconstruction loss ($\mathcal{L}_{\text{MSE}}$), perceptual loss ($\mathcal{L}_{\text{LPIPS}}$), and consistency loss ($\mathcal{L}_{\text{CONSIS}}$), and assess the performance under both novel view and novel pose synthesis tasks. Table~\ref{tab:6} reports the quantitative comparison across different weight settings. The results show that the model achieves the best performance across all metrics when all three loss weights are set to 1.0, confirming the effectiveness of our joint optimization strategy proposed in the loss design. In particular, for the novel pose synthesis task, appropriately emphasizing the consistency loss contributes to improved structural continuity, reducing artifacts such as limb discontinuities and texture drift, thereby enabling more stable human modeling and rendering performance.

\section{Limitations}
Our method relies on accurate SMPL pose and shape parameters as modeling priors. However, in real-world scenarios, accurately obtaining these parameters remains challenging due to occlusion, lighting variations, and clothing interference. This reliance limits the applicability of the method in in-the-wild environments. To address this issue, a strategy similar to that in A-NeRF~\cite{su2021nerf} can be adopted, where the SMPL pose parameters are treated as optimizable variables and jointly optimized with the neural rendering network during training. This approach not only helps correct initial estimation errors but also improves the robustness of pose modeling and the final rendering quality, thereby enhancing the method’s adaptability to complex real-world environments.

\section{Conclusion}
In this paper, we propose \textbf{MonoPartNeRF}, a neural representation framework for high-quality dynamic human rendering under monocular viewpoints. By decomposing the human body into multiple semantically consistent structural parts and introducing a bidirectional deformation mechanism with spatiotemporal parameterization for each part, our method significantly improves the naturalness of joint connections and geometric consistency. We construct a unified surface-time representation in the canonical space, effectively alleviating structural distortions and texture drifting caused by non-rigid deformations. Furthermore, by integrating part-level pose embeddings with appearance feature fusion, our approach enhances detail recovery and reconstruction robustness in regions with missing information and occlusions.

Extensive experiments on the ZJU-MoCap and MonoCap datasets demonstrate that MonoPartNeRF outperforms existing state-of-the-art methods in both novel view synthesis and novel pose synthesis tasks, achieving superior performance in terms of structural consistency, perceptual realism, and pixel-level accuracy. Notably, our method exhibits stronger generalization ability and visual stability under monocular input conditions. Ablation studies further validate the effectiveness of key components, including the bidirectional consistency constraints, the pose and appearance fusion strategy, and the UVT-based spatiotemporal modeling approach, in boosting overall performance. MonoPartNeRF offers an efficient, flexible, and extensible solution for monocular dynamic human modeling and high-quality cross-view and cross-pose rendering, demonstrating great potential for practical applications such as virtual human generation, immersive content creation, and remote interaction.





\bibliographystyle{elsarticle-num} 
\bibliography{main}

\begin{thebibliography}{10}
\expandafter\ifx\csname url\endcsname\relax
  \def\url#1{\texttt{#1}}\fi
\expandafter\ifx\csname urlprefix\endcsname\relax\def\urlprefix{URL }\fi
\expandafter\ifx\csname href\endcsname\relax
  \def\href#1#2{#2} \def\path#1{#1}\fi

\bibitem{debevec2000acquiring}
P.~Debevec, T.~Hawkins, C.~Tchou, H.-P. Duiker, W.~Sarokin, M.~Sagar, Acquiring the reflectance field of a human face, in: Proceedings of the 27th annual conference on Computer graphics and interactive techniques, 2000, pp. 145--156.

\bibitem{guo2019relightables}
K.~Guo, P.~Lincoln, P.~Davidson, J.~Busch, X.~Yu, M.~Whalen, G.~Harvey, S.~Orts-Escolano, R.~Pandey, J.~Dourgarian, et~al., The relightables: Volumetric performance capture of humans with realistic relighting, ACM Transactions on Graphics (ToG) 38~(6) (2019) 1--19.

\bibitem{collet2015high}
A.~Collet, M.~Chuang, P.~Sweeney, D.~Gillett, D.~Evseev, D.~Calabrese, H.~Hoppe, A.~Kirk, S.~Sullivan, High-quality streamable free-viewpoint video, ACM Transactions on Graphics (ToG) 34~(4) (2015) 1--13.

\bibitem{dou2016fusion4d}
M.~Dou, S.~Khamis, Y.~Degtyarev, P.~Davidson, S.~R. Fanello, A.~Kowdle, S.~O. Escolano, C.~Rhemann, D.~Kim, J.~Taylor, et~al., Fusion4d: Real-time performance capture of challenging scenes, ACM Transactions on Graphics (ToG) 35~(4) (2016) 1--13.

\bibitem{mildenhall2021nerf}
B.~Mildenhall, P.~P. Srinivasan, M.~Tancik, J.~T. Barron, R.~Ramamoorthi, R.~Ng, Nerf: Representing scenes as neural radiance fields for view synthesis, Communications of the ACM 65~(1) (2021) 99--106.

\bibitem{pons2021d}
G.~Pons-Moll, F.~Moreno-Noguer, E.~Corona, A.~Pumarola, D-nerf: Neural radiance fields for dynamic scenes, in: 2021 IEEE/CVF Conference on Computer Vision and Pattern Recognition (CVPR), IEEE, 2021.

\bibitem{li2021neural}
Z.~Li, S.~Niklaus, N.~Snavely, O.~Wang, Neural scene flow fields for space-time view synthesis of dynamic scenes, in: Proceedings of the IEEE/CVF Conference on Computer Vision and Pattern Recognition, 2021, pp. 6498--6508.

\bibitem{peng2021animatable}
S.~Peng, J.~Dong, Q.~Wang, S.~Zhang, Q.~Shuai, X.~Zhou, H.~Bao, Animatable neural radiance fields for modeling dynamic human bodies, in: Proceedings of the IEEE/CVF International Conference on Computer Vision, 2021, pp. 14314--14323.

\bibitem{weng2022humannerf}
C.-Y. Weng, B.~Curless, P.~P. Srinivasan, J.~T. Barron, I.~Kemelmacher-Shlizerman, Humannerf: Free-viewpoint rendering of moving people from monocular video, in: Proceedings of the IEEE/CVF conference on computer vision and pattern Recognition, 2022, pp. 16210--16220.

\bibitem{yu2023monohuman}
Z.~Yu, W.~Cheng, X.~Liu, W.~Wu, K.-Y. Lin, Monohuman: Animatable human neural field from monocular video, in: Proceedings of the IEEE/CVF Conference on Computer Vision and Pattern Recognition, 2023, pp. 16943--16953.

\bibitem{li2023posevocab}
Z.~Li, Z.~Zheng, Y.~Liu, B.~Zhou, Y.~Liu, Posevocab: Learning joint-structured pose embeddings for human avatar modeling, in: ACM SIGGRAPH 2023 conference proceedings, 2023, pp. 1--11.

\bibitem{geng2023learning}
C.~Geng, S.~Peng, Z.~Xu, H.~Bao, X.~Zhou, Learning neural volumetric representations of dynamic humans in minutes, in: Proceedings of the IEEE/CVF Conference on Computer Vision and Pattern Recognition, 2023, pp. 8759--8770.

\bibitem{schonberger2016structure}
J.~L. Schonberger, J.-M. Frahm, Structure-from-motion revisited, in: Proceedings of the IEEE conference on computer vision and pattern recognition, 2016, pp. 4104--4113.

\bibitem{schonberger2016pixelwise}
J.~L. Sch{\"o}nberger, E.~Zheng, J.-M. Frahm, M.~Pollefeys, Pixelwise view selection for unstructured multi-view stereo, in: Computer Vision--ECCV 2016: 14th European Conference, Amsterdam, The Netherlands, October 11-14, 2016, Proceedings, Part III 14, Springer, 2016, pp. 501--518.

\bibitem{su2020robustfusion}
Z.~Su, L.~Xu, Z.~Zheng, T.~Yu, Y.~Liu, L.~Fang, Robustfusion: Human volumetric capture with data-driven visual cues using a rgbd camera, in: Computer Vision--ECCV 2020: 16th European Conference, Glasgow, UK, August 23--28, 2020, Proceedings, Part IV 16, Springer, 2020, pp. 246--264.

\bibitem{saito2019pifu}
S.~Saito, Z.~Huang, R.~Natsume, S.~Morishima, A.~Kanazawa, H.~Li, Pifu: Pixel-aligned implicit function for high-resolution clothed human digitization, in: Proceedings of the IEEE/CVF international conference on computer vision, 2019, pp. 2304--2314.

\bibitem{saito2020pifuhd}
S.~Saito, T.~Simon, J.~Saragih, H.~Joo, Pifuhd: Multi-level pixel-aligned implicit function for high-resolution 3d human digitization, in: Proceedings of the IEEE/CVF conference on computer vision and pattern recognition, 2020, pp. 84--93.

\bibitem{chen2021snarf}
X.~Chen, Y.~Zheng, M.~J. Black, O.~Hilliges, A.~Geiger, Snarf: Differentiable forward skinning for animating non-rigid neural implicit shapes, in: Proceedings of the IEEE/CVF International Conference on Computer Vision, 2021, pp. 11594--11604.

\bibitem{mihajlovic2022coap}
M.~Mihajlovic, S.~Saito, A.~Bansal, M.~Zollhoefer, S.~Tang, Coap: Compositional articulated occupancy of people, in: Proceedings of the IEEE/CVF Conference on Computer Vision and Pattern Recognition, 2022, pp. 13201--13210.

\bibitem{saito2021scanimate}
S.~Saito, J.~Yang, Q.~Ma, M.~J. Black, Scanimate: Weakly supervised learning of skinned clothed avatar networks, in: Proceedings of the IEEE/CVF Conference on Computer Vision and Pattern Recognition, 2021, pp. 2886--2897.

\bibitem{kwon2021neural}
Y.~Kwon, D.~Kim, D.~Ceylan, H.~Fuchs, Neural human performer: Learning generalizable radiance fields for human performance rendering, Advances in Neural Information Processing Systems 34 (2021) 24741--24752.

\bibitem{liu2021neural}
L.~Liu, M.~Habermann, V.~Rudnev, K.~Sarkar, J.~Gu, C.~Theobalt, Neural actor: Neural free-view synthesis of human actors with pose control, ACM transactions on graphics (TOG) 40~(6) (2021) 1--16.

\bibitem{remelli2022drivable}
E.~Remelli, T.~Bagautdinov, S.~Saito, C.~Wu, T.~Simon, S.-E. Wei, K.~Guo, Z.~Cao, F.~Prada, J.~Saragih, et~al., Drivable volumetric avatars using texel-aligned features, in: ACM SIGGRAPH 2022 conference proceedings, 2022, pp. 1--9.

\bibitem{chen2021mvsnerf}
A.~Chen, Z.~Xu, F.~Zhao, X.~Zhang, F.~Xiang, J.~Yu, H.~Su, Mvsnerf: Fast generalizable radiance field reconstruction from multi-view stereo, in: Proceedings of the IEEE/CVF international conference on computer vision, 2021, pp. 14124--14133.

\bibitem{johari2022geonerf}
M.~M. Johari, Y.~Lepoittevin, F.~Fleuret, Geonerf: Generalizing nerf with geometry priors, in: Proceedings of the IEEE/CVF Conference on Computer Vision and Pattern Recognition, 2022, pp. 18365--18375.

\bibitem{wang2021ibrnet}
Q.~Wang, Z.~Wang, K.~Genova, P.~P. Srinivasan, H.~Zhou, J.~T. Barron, R.~Martin-Brualla, N.~Snavely, T.~Funkhouser, Ibrnet: Learning multi-view image-based rendering, in: Proceedings of the IEEE/CVF conference on computer vision and pattern recognition, 2021, pp. 4690--4699.

\bibitem{chen2019learning}
Z.~Chen, H.~Zhang, Learning implicit fields for generative shape modeling, in: Proceedings of the IEEE/CVF conference on computer vision and pattern recognition, 2019, pp. 5939--5948.

\bibitem{liu2020dist}
S.~Liu, Y.~Zhang, S.~Peng, B.~Shi, M.~Pollefeys, Z.~Cui, Dist: Rendering deep implicit signed distance function with differentiable sphere tracing, in: Proceedings of the IEEE/CVF Conference on Computer Vision and Pattern Recognition, 2020, pp. 2019--2028.

\bibitem{park2019deepsdf}
J.~J. Park, P.~Florence, J.~Straub, R.~Newcombe, S.~Lovegrove, Deepsdf: Learning continuous signed distance functions for shape representation, in: Proceedings of the IEEE/CVF conference on computer vision and pattern recognition, 2019, pp. 165--174.

\bibitem{sitzmann2019scene}
V.~Sitzmann, M.~Zollh{\"o}fer, G.~Wetzstein, Scene representation networks: Continuous 3d-structure-aware neural scene representations, Advances in neural information processing systems 32 (2019).

\bibitem{zhi2021place}
S.~Zhi, T.~Laidlow, S.~Leutenegger, A.~J. Davison, In-place scene labelling and understanding with implicit scene representation, in: Proceedings of the IEEE/CVF International Conference on Computer Vision, 2021, pp. 15838--15847.

\bibitem{alldieck2022photorealistic}
T.~Alldieck, M.~Zanfir, C.~Sminchisescu, Photorealistic monocular 3d reconstruction of humans wearing clothing, in: Proceedings of the IEEE/CVF Conference on Computer Vision and Pattern Recognition, 2022, pp. 1506--1515.

\bibitem{hu2023sherf}
S.~Hu, F.~Hong, L.~Pan, H.~Mei, L.~Yang, Z.~Liu, Sherf: Generalizable human nerf from a single image, in: Proceedings of the IEEE/CVF International Conference on Computer Vision, 2023, pp. 9352--9364.

\bibitem{loper2023smpl}
M.~Loper, N.~Mahmood, J.~Romero, G.~Pons-Moll, M.~J. Black, Smpl: A skinned multi-person linear model, in: Seminal Graphics Papers: Pushing the Boundaries, Volume 2, 2023, pp. 851--866.

\bibitem{jiang2023instantavatar}
T.~Jiang, X.~Chen, J.~Song, O.~Hilliges, Instantavatar: Learning avatars from monocular video in 60 seconds, in: Proceedings of the IEEE/CVF Conference on Computer Vision and Pattern Recognition, 2023, pp. 16922--16932.

\bibitem{xu2021h}
H.~Xu, T.~Alldieck, C.~Sminchisescu, H-nerf: Neural radiance fields for rendering and temporal reconstruction of humans in motion, Advances in Neural Information Processing Systems 34 (2021) 14955--14966.

\bibitem{jiang2022neuman}
W.~Jiang, K.~M. Yi, G.~Samei, O.~Tuzel, A.~Ranjan, Neuman: Neural human radiance field from a single video, in: European Conference on Computer Vision, Springer, 2022, pp. 402--418.

\bibitem{peng2023implicit}
S.~Peng, C.~Geng, Y.~Zhang, Y.~Xu, Q.~Wang, Q.~Shuai, X.~Zhou, H.~Bao, Implicit neural representations with structured latent codes for human body modeling, IEEE Transactions on Pattern Analysis and Machine Intelligence 45~(8) (2023) 9895--9907.

\bibitem{su2021nerf}
S.-Y. Su, F.~Yu, M.~Zollh{\"o}fer, H.~Rhodin, A-nerf: Articulated neural radiance fields for learning human shape, appearance, and pose, Advances in neural information processing systems 34 (2021) 12278--12291.

\bibitem{hu2022hvtr}
T.~Hu, T.~Yu, Z.~Zheng, H.~Zhang, Y.~Liu, M.~Zwicker, Hvtr: Hybrid volumetric-textural rendering for human avatars, in: 2022 International Conference on 3D Vision (3DV), IEEE, 2022, pp. 197--208.

\bibitem{jiang2022selfrecon}
B.~Jiang, Y.~Hong, H.~Bao, J.~Zhang, Selfrecon: Self reconstruction your digital avatar from monocular video, in: Proceedings of the IEEE/CVF Conference on Computer Vision and Pattern Recognition, 2022, pp. 5605--5615.

\bibitem{lewis2023pose}
J.~P. Lewis, M.~Cordner, N.~Fong, Pose space deformation: a unified approach to shape interpolation and skeleton-driven deformation, in: Seminal Graphics Papers: Pushing the Boundaries, Volume 2, 2023, pp. 811--818.

\bibitem{deng2020nasa}
B.~Deng, J.~P. Lewis, T.~Jeruzalski, G.~Pons-Moll, G.~Hinton, M.~Norouzi, A.~Tagliasacchi, Nasa neural articulated shape approximation, in: Computer Vision--ECCV 2020: 16th European Conference, Glasgow, UK, August 23--28, 2020, Proceedings, Part VII 16, Springer, 2020, pp. 612--628.

\bibitem{johnson2016perceptual}
J.~Johnson, A.~Alahi, L.~Fei-Fei, Perceptual losses for real-time style transfer and super-resolution, in: Computer Vision--ECCV 2016: 14th European Conference, Amsterdam, The Netherlands, October 11-14, 2016, Proceedings, Part II 14, Springer, 2016, pp. 694--711.

\bibitem{peng2024animatable}
S.~Peng, Z.~Xu, J.~Dong, Q.~Wang, S.~Zhang, Q.~Shuai, H.~Bao, X.~Zhou, Animatable implicit neural representations for creating realistic avatars from videos, TPAMI (2024).

\bibitem{habermann2021real}
M.~Habermann, L.~Liu, W.~Xu, M.~Zollhoefer, G.~Pons-Moll, C.~Theobalt, Real-time deep dynamic characters, ACM Transactions on Graphics (ToG) 40~(4) (2021) 1--16.

\bibitem{habermann2020deepcap}
M.~Habermann, W.~Xu, M.~Zollhofer, G.~Pons-Moll, C.~Theobalt, Deepcap: Monocular human performance capture using weak supervision, in: Proceedings of the IEEE/CVF Conference on Computer Vision and Pattern Recognition, 2020, pp. 5052--5063.

\bibitem{peng2021neural}
S.~Peng, Y.~Zhang, Y.~Xu, Q.~Wang, Q.~Shuai, H.~Bao, X.~Zhou, Neural body: Implicit neural representations with structured latent codes for novel view synthesis of dynamic humans, in: Proceedings of the IEEE/CVF Conference on Computer Vision and Pattern Recognition, 2021, pp. 9054--9063.

\bibitem{hu2024gauhuman}
S.~Hu, T.~Hu, Z.~Liu, Gauhuman: Articulated gaussian splatting from monocular human videos, in: Proceedings of the IEEE/CVF conference on computer vision and pattern recognition, 2024, pp. 20418--20431.

\bibitem{paszke2019pytorch}
A.~Paszke, Pytorch: An imperative style, high-performance deep learning library, arXiv preprint arXiv:1912.01703 (2019).

\bibitem{kingma2014adam}
D.~P. Kingma, J.~Ba, Adam: A method for stochastic optimization, arXiv preprint arXiv:1412.6980 (2014).

\end{thebibliography}







\end{document}